\documentclass[runningheads]{llncs}

\usepackage{eccv}

\usepackage{eccvabbrv}

\usepackage{graphicx}
\usepackage{booktabs}

\usepackage[accsupp]{axessibility}  

\usepackage{multirow}

\definecolor{IKKONZOME}{RGB}{244,167,185} 
\definecolor{HIWAMOEGI}{RGB}{144,180,75} 
\definecolor{YAMABUKI}{RGB}{255,177,27} 

\usepackage{soul} 
\usepackage{xcolor} 
\usepackage{tcolorbox}
\usepackage{makecell}
\usepackage{wrapfig}

\usepackage{hyperref}

\usepackage{orcidlink}

\begin{document}

\sethlcolor{HIWAMOEGI!40} 

\title{AgriBench: A Hierarchical Agriculture Benchmark for Multimodal Large Language Models} 

\titlerunning{AgriBench}

\author{Yutong Zhou\inst{1}\orcidlink{0000-0001-5018-3501} \and
Masahiro Ryo\inst{1,2}\orcidlink{0000-0002-5271-3446}}

\authorrunning{Y. Zhou and M. Ryo}

\institute{Leibniz Centre for Agricultural Landscape Research (ZALF), Eberswalder Str. 84, 15374, Müncheberg, Germany
\email{yutong.zhou@zalf.de}\\ \and
Brandenburg University of Technology Cottbus–Senftenberg, Platz Der Deutschen Einheit 1, 03046, Cottbus, Germany}


\maketitle

\centerline{\href{https://github.com/Yutong-Zhou-cv/AgriBench}{\texttt{https://github.com/Yutong-Zhou-cv/AgriBench}}}

\begin{abstract}
We introduce AgriBench, the first agriculture benchmark designed to evaluate MultiModal Large Language Models (MM-LLMs) for agriculture applications.
To further address the agriculture knowledge-based dataset limitation problem, we propose MM-LUCAS, a multimodal agriculture dataset, that includes 1,784 landscape images, segmentation masks, depth maps, and detailed annotations (geographical location, country, date, land cover and land use taxonomic details, quality scores, aesthetic scores, \etc), based on the Land Use/Cover Area Frame Survey (LUCAS) dataset, which contains comparable statistics on land use and land cover for the European Union (EU) territory. 
This work presents a groundbreaking perspective in advancing agriculture MM-LLMs and is still in progress, offering valuable insights for future developments and innovations in specific expert knowledge-based MM-LLMs. 

  \keywords{Agriculture Benchmark \and Dataset \and Hierarchical Evaluations}
\end{abstract}

\section{Introduction} 
\label{sec:intro}

\begin{quotation}
\textit{``Agriculture is the most healthful, most useful and most noble employment of man.''}

\rightline{-- George Washington (1732–1799)}
\end{quotation}


Agriculture is an important foundation for human existence. Nearly half of the terrestrial land is used for agriculture in Europe, contributing to food, fiber, and bio-energy resource production\cite{maes2013mapping}. Agriculture depends on agroecosystems comprising subterranean soil, soil organisms, habitats for wild flora, and animals in and around the fields. Traditional agricultural practices heavily depended on the empirical knowledge and expertise of farmers to achieve productive yields. To increase efficiency and optimize production, digitalization has been an important agenda in agriculture for increasing efficiency and optimization, including the utility of Artificial Intelligence (AI).  AI has rapidly developed and is widely used to promote agricultural automation. This advancement has significant promise for enhancing agricultural processes, including efficient assessment, explanation, informed decision-making, and understanding of agricultural systems. 

There have been several significant advancements in Machine Learning (ML) and Deep Learning (DL) domains within the agriculture and biodiversity research field in the past decade, such as species identification\cite{kong2021multi}, wildlife monitoring and protection\cite{roy2023wildect}, plant disease detection\cite{kotwal2023agricultural,wang2024vegetable}, plant phenotyping\cite{li2020review}, crop classification\cite{teixeira2023deep}, weed detection\cite{ong2023uav}, intelligent spraying\cite{hafeez2023implementation,seol2022field}, robotic harvesting\cite{tang2020recognition,kok2024occluded}, \etc. Despite these achievements, conventional ML and DL models still have certain limitations: require extensive, task-specific, and well-labeled datasets for effective training; only adapt to specific tasks but cannot generalize to other tasks or unseen data. Due to these limitations, several approaches have been examined, including transfer learning\cite{nayak2023application}, few-shot learning\cite{rezaei2024plant}, label-efficient learning\cite{li2023label}, self-supervised learning\cite{zhao2023cla}, to name a few.  

Recently, benefiting from the advancements in Large Language Models (LLMs)\cite{achiam2023gpt,bai2023qwen,touvron2023llama}, MultiModal Large Language Models (MM-LLMs) have rapidly become a new paradigm bridging the fields of Natural Language Processing (NLP) and Computer Vision (CV). MM-LLMs preserve the inherent reasoning and decision-making capabilities of LLMs and showcase remarkable versatility and efficiency across a diverse range of multimodal (MM) tasks\cite{zhang2024mm}, such as emotional understanding\cite{yang2024emollm}, image captioning\cite{rotstein2024fusecap,li2024if}, Visual Question Answering (VQA)\cite{hu2024prompting}, \etc. However, the agricultural domain involves more complex and expert tasks, including multimodality and human-nature interactions, which present significant challenges even for advanced MM-LLMs. Therefore, it is essential to develop benchmarks to evaluate the performance of the existing models specialized for agriculture. To our knowledge, no such MM-LLM benchmarks currently exist.

Here, we present four key questions about MM-LLMs in the agriculture domain, which will be discussed in detail in the following sections.

\textbf{Q1:} \textit{How to evaluate the MM-LLM's capacities in agriculture?} 

\textbf{A1: [\cref{sec:agribench}]} Due to the unique needs and complexity of agricultural research, it is important to build an agriculture-specified benchmark to evaluate the MM-LLMs. Thus, we propose a novel \textbf{Agri}culture \textbf{Bench}mark: \textbf{AgriBench}.

\textbf{Q2:} \textit{How to design an agriculture multimodal dataset? } 

\textbf{A2: [\cref{sec:data}]} To address the lack of agricultural datasets, we create \textbf{MM-LUCAS} with 1,784 annotated agricultural scenery images from 27 EU countries.

\textbf{Q3:} \textit{How effective are advanced MM-LLMs in solving agricultural problems without extra fine-tuning?} 

\textbf{A3: [\cref{sec:exp}]} MM-LLMs understand general agricultural content well but struggle with specific problems like diagnosing plant diseases without fine-tuning.





Our contributions can be summarized as follows:
\begin{enumerate}
    \item AgriBench is the first hierarchical benchmark to evaluate the comprehension and reasoning abilities of existing MM-LLMs regarding agriculture.
    \item An innovative MM-LUCAS dataset is also newly designed for the proposed AgriBench, which contains corresponding multi-modality annotations.
    \item We evaluate the capability of 5 MM-LLMs on our benchmark and present some promising directions for future expert knowledge-based MM-LLMs.
\end{enumerate}

\section{Related Works}
\label{sec:review}

\subsubsection{Multimodal Large Language Models (MM-LLMs)}
\label{sec:review1}
The rapid developments and remarkable achievements of LLMs have driven a growing research interest in MM-LLMs. These models enhance multimodal comprehension by aligning visual features from pre-trained image encoders with LLMs on image-text datasets. Groundbreaking MM-LLMs, such as Flamingo\cite{alayrac2022flamingo}, GPT-4\cite{achiam2023gpt}, ModaVers\cite{wang2024modaverse}, Cambrian-1\cite{tong2024cambrian} and LLaVA-OneVision\cite{li2024llava} have successfully fused visual data and text and adapted to various multimodal tasks.
However, regarding specialized research fields and reality applications, such as industry, agriculture, and healthcare, existing advanced MM-LLMs still face significant challenges in accurately and comprehensively handling domain-specific tasks. 

\subsubsection{Benchmarks for Multimodal Large Language Models}
\label{sec:review2}
Current benchmarks mainly focus on the ability to predict the understanding of vision-text inputs. For instance, MMBench\cite{liu2023mmbench} creates extensive question sets to enhance the objective evaluation of MM-LLMs. GVTBench\cite{wang2023makes} is designed for two new tasks for MM-LLMs (object counting and multi-class identification), however, its evaluations are limited to some specific aspects of visual understanding. For specialized domains, evaluation requires a level of expertise and precision that current benchmarks often fail to achieve. Therefore, it is essential to develop MM-LLM benchmarks specifically customized for these research domains.

\subsubsection{Land Use/Cover Area Frame Survey (LUCAS)}
\label{sec:review3}

The European Union (EU) encompasses a wide variety of landscapes and ecosystems, from densely populated urban centers to sparsely inhabited rural areas.
Knowing the patterns in ``Land Cover'' and ``Land Use'' is important for human activity and geography. 

The Land Use/Cover Area Frame Survey (LUCAS) is one of the most extensive and authoritative in-situ field surveys across Europe. Land cover (LC) denotes the visible physical and biological features, such as cropland and waterbody. Land use (LU) refers to how humans utilize the land for socio-economic purposes, such as residential living and agriculture\cite{web:eurostat}. 
From 2006 to 2018, LUCAS was conducted every three years across EU member states, providing a standardized framework for collecting statistics on LC and LU and other information such as soil physico-chemical parameters. Data was collected from 1,351,293 points at 651,780 unique locations, covering 106 variables, and including 5.4 million scenery photos\cite{d2020harmonised}. 
The LUCAS data includes (1) Microdata on LC, LU, and environmental parameters; (2) Landscape images from the four cardinal directions (north, south, east, and west); and (3) Statistical tables aggregating estimates of LC and LU at the geographical level based on the microdata.
This survey has significantly contributed to understanding agriculture, environmental conditions, and sustainable development across the EU. Previous studies published several datasets\cite{laso2020crowdsourcing,yordanov2023crop} based on the extensive information of LUCAS. Recently, Martinez-Sanchez \etal published a segmentation dataset\cite{martinez2024semantic}, we use this database to develop a novel MM-LLM benchmark dataset.

\section{AgriBench}
\label{sec:agribench}
We introduce AgriBench, the first hierarchical benchmark designed to assess the visual comprehension capabilities of MM-LLMs in the agricultural domain. Most existing MM-LLM benchmarks mainly focus on multimodal complexity. However, given the unique characteristics and requirements of specific domains, we argue that task complexity must also be taken into account. Multimodal complexity and task complexity should be considered as two distinct but correlated axes. It is feasible to encounter complex tasks involving a single modality or simple tasks requiring multimodal data. To address this, AgriBench designs diverse scenario tasks that reflect real-world agricultural challenges, ensuring a robust assessment of model performance. Furthermore, AgriBench evaluates models across five levels of task complexity and various modalities, providing a comprehensive framework for advancements in agricultural AI.

\begin{figure}[t]
    \centering
    \begin{minipage}{0.53\textwidth}
        \centering
        \includegraphics[width=\textwidth]{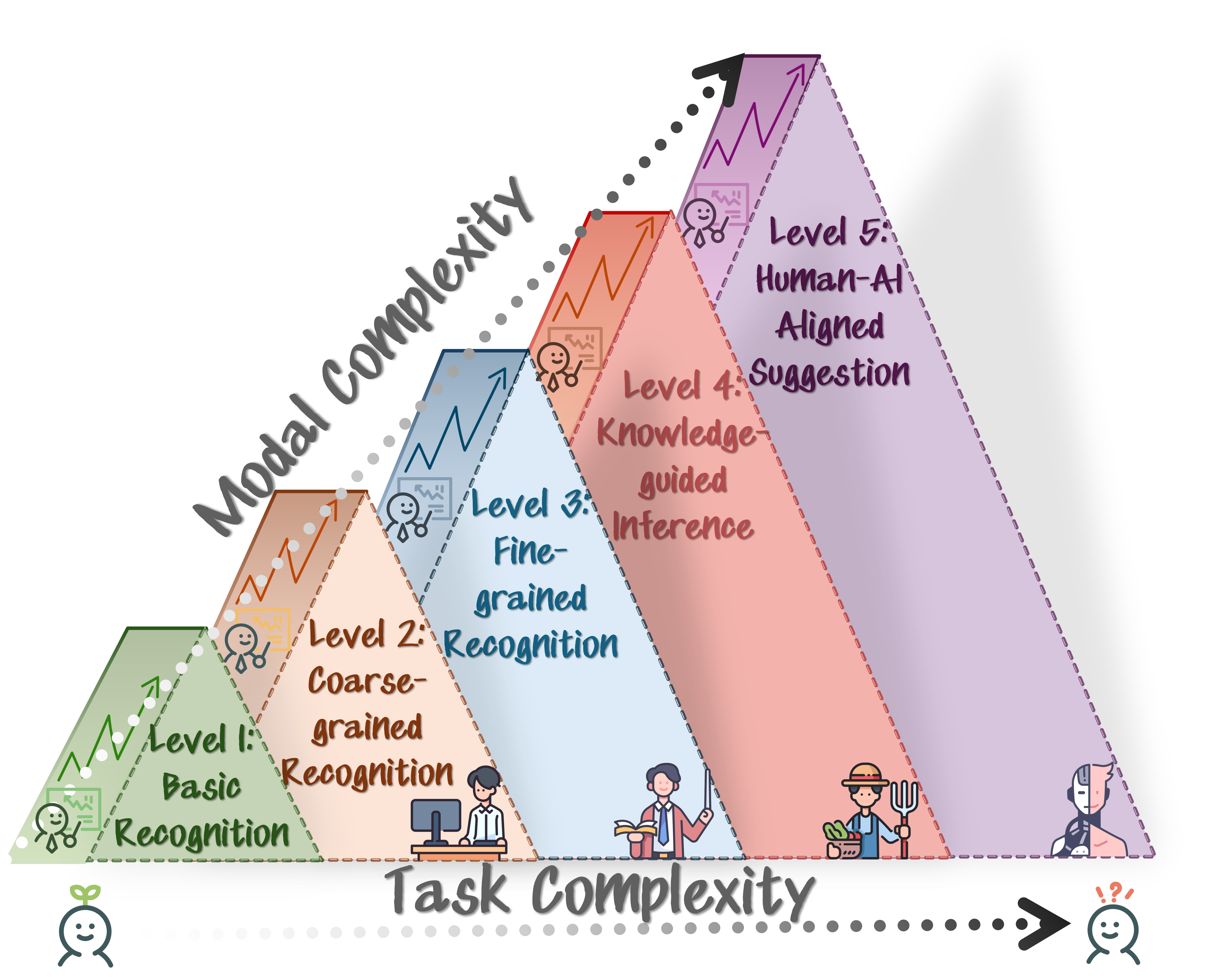} 
        \caption{Five levels of MM task difficulty in the agricultural domain.}
        \label{fig:AgriBench}
    \end{minipage}\hfill
    \begin{minipage}{0.43\textwidth}
        \centering
        \captionof{table}{Abbreviations}
        \fontsize{6.5}{10}\selectfont 
        \label{tab:Abb}
        \begin{tabular}{ll}
\textbf{AgriBench:} & Agriculture Benchmark            \\
\textbf{AI:}        & Artificial Intelligence          \\
\textbf{AS:}        & Aesthetic Score                  \\
\textbf{EU:}        & European Union                   \\
\textbf{I:}         & Image                            \\
\textbf{LC:}        & Land Cover                       \\
\textbf{LU:}        & Land Use                         \\
\textbf{LUCAS:}     & Land Use/Cover Area Frame Survey \\
\textbf{MM:}        & Multi-modal\\
\textbf{MM-LLMs:}   & Multimodal Large Language Models \\
\textbf{QS:}        & Quality Score                    \\
\textbf{SOTA:}      & State-of-the-Art                 \\
\textbf{T:}         & Text
        \end{tabular}
        \label{tab:AgriBench}
    \end{minipage}
\end{figure}



\subsection{Hierarchical Standard}
\label{sec:standard}

Agriculture requires a broad range of tasks, from simple object detection to complex decision-making (\eg, fertilization strategy). In particular, high-stakes decision-making (\eg, planning and management) typically has no objectively single correct answer. There will be several answers equally convincing for domain experts. Therefore, this ambiguity demands to be addressed within the context of human-centered AI.
As illustrated in \cref{fig:AgriBench}, we define the capabilities of MM-LLMs in agriculture into hierarchical levels standards ranging from L1 (Basic Recognition) to L5 (Human-AI Aligned Suggestion). 


\subsection{The 5 Levels of Agriculture MM-LLMs Evaluation Strategy}
\label{sec:Strategy}
We propose the 5 levels of MM-LLM capability that can assess to which extent the model can address multiple modalities and various tasks. For each level, we first describe the definition and some agriculture task examples, ranging from perception to cognition, simple to complex. Then, we summarize totally 17 main tasks with detailed descriptions as follows, with the template: ``\colorbox {IKKONZOME!10}{T$\rightarrow$T}/\colorbox {IKKONZOME!20}{I$\rightarrow$T}/\colorbox {IKKONZOME!30}{I$\rightarrow$I}/\colorbox {IKKONZOME!60}{T+I$\rightarrow$T}/\colorbox {IKKONZOME!90}{T+I$\rightarrow$I} \textit{agriculture tasks:} details.''

\subsubsection{Level 1: Basic Recognition}
\label{sec:l1}

At the lowest level, the model should accurately describe visual content in images. The textual descriptions must be immediately clear to the user based on human intuition, without additional justification to evaluate performance. This includes object detection (\eg, identifying fruits on a tree, flowering plants, and weeds on soil), boundary delineation (\eg, distinguishing soil from plants), and species classification (\eg, categorizing visually distinct common crops such as sunflowers, lavenders, wheat, maize, and cotton).

\begin{itemize}
    \item \colorbox {IKKONZOME!10}{T$\rightarrow$T} \textit{Basic Question Answering}: Describe the basic information on crops or scenes based on broad knowledge from LLMs without complex inference. 
    \item \colorbox {IKKONZOME!60}{T+I$\rightarrow$T} \textit{Species Classification}: Detect and classify common crops. 
    \item \colorbox {IKKONZOME!60}{T+I$\rightarrow$T} \textit{Image Captioning}: Describe the visible agriculture objects.
\end{itemize}

\subsubsection{Level 2: Coarse-grained Recognition}
\label{sec:l2}

The model can describe clearly definable properties. These properties should be objectively extractable from the MM-LLMs, although the extraction process may require time for the user. While the task complexity is higher than Level 1, no inference is still necessary, and users should consistently agree on the results. This includes object counting (\eg, counting the number of fruits in an image or on a specific branch), dominant class detection (\eg, identifying the most prevalent crop type in the image if multiple crops are seen), and classification (\eg, predicting widely recognized basic phenological growth stages such as seedling, vegetative, flowering, and ripening).

\begin{itemize}
    \item \colorbox {IKKONZOME!60}{T+I$\rightarrow$T} \textit{Object counting}: Count the number of user-specified objects.
    \item \colorbox {IKKONZOME!60}{T+I$\rightarrow$T} \textit{Basic Scene Analysis}: Generate detailed textual descriptions of multiple visible agriculture objects and the overall scene accurately.
\end{itemize}

\subsubsection{Level 3: Fine-grained Recognition}
\label{sec:l3}

The model at this level can describe fine-grained properties, though the estimation involves some subjectivity. Unlike Level 2, while a common consensus on the answer can be reached, there may still be some disagreement. This includes tasks such as image enhancement (\eg, multiple enhancement methods are possible) and grounded image captioning (\eg, the explanation focus could be various if multiple objects are present).

\begin{itemize}
    \item \colorbox {IKKONZOME!10}{T$\rightarrow$T} \textit{Advanced Question Answering}: Describe the in-depth, detailed information on crops or scenes based on specific knowledge from existing LLMs that may require some degree of inference. 
    \item \colorbox {IKKONZOME!60}{T+I$\rightarrow$T} \textit{Dense object counting}: Counting dense objects such as the numbers of individual plants, cherries, and flowers (Note that the number can be more than 50-100 in a single image). 
    \item \colorbox {IKKONZOME!60}{T+I$\rightarrow$T} \textit{Contextual Scene Analysis}: Generate fine-grained descriptions of scenes considering various agricultural objects. Furthermore, evaluate the degree of aesthetics based on the analysis. 
    \item \colorbox {IKKONZOME!90}{T+I$\rightarrow$I} \textit{Image Enhancement}: Recover high-quality images from low-quality (extreme weather, motion blur, or low light) images. 
\end{itemize}

\subsubsection{Level 4: Knowledge-guided Inference}
\label{sec:l4}

The model can describe elements that are not directly visible, requiring educated guesses based on expert insights or domain-specific knowledge. Experienced users can make informed guesses using visible properties as hints, often relying on empirical correlations with a high degree of flexibility. Thus, adding additional data sources can enhance the accuracy of these predictions. This includes tasks such as regression (\eg, estimating crop yield, plant health, soil health), classification (\eg, identifying plant disease, crop phenotype, and geographic region), and scene generation.

\begin{itemize}
    \item \colorbox {IKKONZOME!20}{I$\rightarrow$T} \textit{Surrounding Scene Analysis}: Predict crop yield based on visible properties (plant size, leaf color, and density) with historical data and weathers. 
    \item \colorbox {IKKONZOME!30}{I$\rightarrow$I} \textit{Automatic Image Enhancement}: Automatically enhance images from a comprehensive scene analysis, without relying on additional guidance.  
    \item \colorbox {IKKONZOME!60}{T+I$\rightarrow$T} \textit{Environmental Impact Prediction}: Combine visual images with information on fertilizer use, irrigation practices, and local biodiversity to predict the environmental impact.
    \item \colorbox {IKKONZOME!60}{T+I$\rightarrow$T} \textit{Camouflaged Object Detection}: Identify camouflaged objects blended with surroundings, aiding in pest activity risk assessment.
    \item \colorbox {IKKONZOME!90}{T+I$\rightarrow$I} \textit{Scene Generation}: Generate scenes for future scenarios based on current observations.

\end{itemize}

\subsubsection{Level 5: Human Aligned Suggestion}
\label{sec:l5}
Level 5 extends beyond inference based on expert insights. The model should be capable of suggesting actions or scenarios for future implementation. These suggestions could influence high-stakes decision-making, requiring the model to provide comprehensive justifications, which ensures that users can assess the validity and feasibility of the suggestions before taking action.

\begin{itemize}
    \item \colorbox {IKKONZOME!10}{T$\rightarrow$T} \textit{Strategic Planning}: Suggest long-term strategies using continuously updated current and historical data, such as recommending crop rotation and irrigation schedules to maximize yield and soil health over multiple seasons. 
    \item \colorbox {IKKONZOME!20}{I$\rightarrow$T} \textit{Scene Projection}: Evaluate various ``what-if'' scenarios, such as the impact of changing a particular farming practice or responding to an unexpected event (e.g., sudden weather changes), and suggest the best course of action with a detailed analysis of potential outcomes and probabilities.
    \item \colorbox {IKKONZOME!60}{T+I$\rightarrow$T} \textit{Sustainability Recommendations}: Propose sustainable farming practices that balance productivity with environmental conservation, such as reducing chemical use, or adopting no-till farming, based on accurate prediction and detailed analysis results.
\end{itemize}

\section{MM-LUCAS Dataset}
\label{sec:data}
Image acquisition and computer vision play a central role in advancing agricultural digitalization and optimization. Images are commonly captured using handheld cameras\cite{han2021real}, Unmanned Aerial Vehicles (UAV)-mounted cameras\cite{herrmann2020assessment}, and earth observatories\cite{tziolas2021earth}. While multispectral (\eg, near-infrared) and hyperspectral imaging techniques are increasingly popular for estimating plant and soil conditions across various wavelengths, RGB images still serve as the basis for various agricultural applications.

\begin{figure}[t]
    \centering
    \includegraphics[width=1\linewidth]{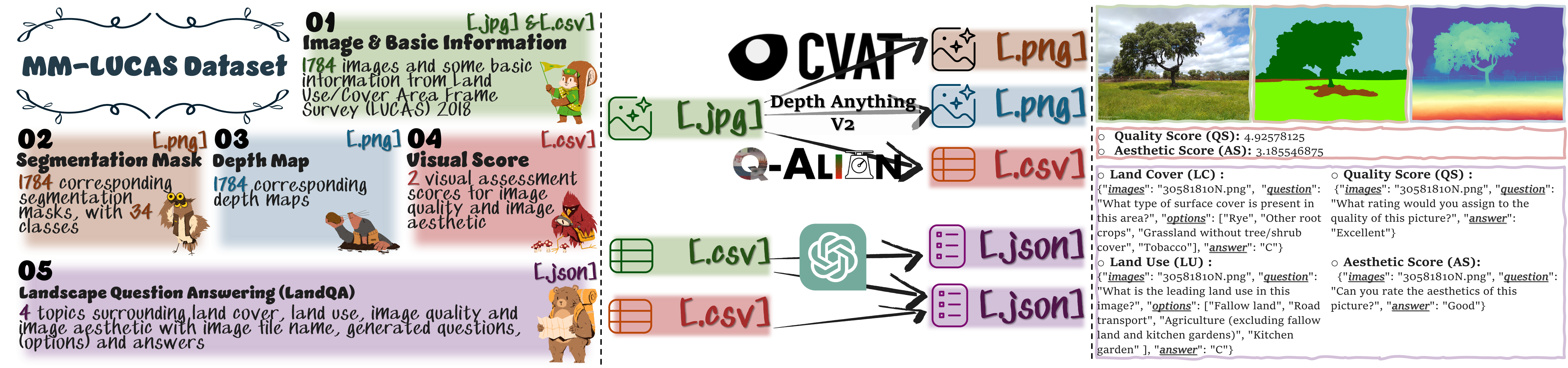}
    \caption{\textit{Left}: Overview of the MM-LUCAS dataset. \textit{Middle}: The illustration of data collection and processing. \textit{Right}: Data example in the MM-LUCAS dataset. \textit{The box colors in the Middle and Right correspond to the colors in the Left.}}
    \label{fig:data-overview}
\end{figure}

However, there is a lack of publicly available and well-annotated agricultural image datasets with multimodal information. To address this gap and enhance the agricultural knowledge understanding capabilities of MM-LLMs, we propose a novel multimodal agriculture dataset with specialized annotations, designed to follow the proposed AgriBench, named MM-LUCAS. As shown in \cref{fig:data-overview}, our dataset has the following properties: (1) It contains 1,784 scenery images and some basic information from the original LUCAS dataset. (2) It includes 1,784 corresponding semantic segmentation masks and depth maps. (3) For each image, we assess the quality and aesthetic. (4) We generate landscape question answering with 4 topics. MM-LUCAS was developed based on the existing segmentation LUCAS dataset\cite{martinez2024semantic}.
We describe the detailed information below.

\begin{table}[]
\centering
\caption{Detail of microdata in MM-LUCAS dataset.}
\fontsize{6.5}{8}\selectfont 
\begin{tabular}{lll||c}
\hline
\textbf{Source}           & \textbf{Label}   & \textbf{{\tiny{{[}Data Type{]}}} Description}       & \textbf{Example}                                        \\ \hline
\multirow{9}{*}{\cite{martinez2024semantic}}      & file             & {[}String{]} Image file name               & 30581810N.png                                           \\
                          & gps\_long        & {\tiny{{[}Double{]}}} GPS observation longitude     & -4.434544                                               \\
                          & gps\_lat         & {\tiny{{[}Double{]}}} GPS observation latitude      & 38.29408                                                \\
                          & date             & {\tiny{{[}Date{]}}} Date of observation             & 04/06/2018  12:00:00                                    \\
                          & nuts0            & {\tiny{{[}String{]}}} NUTS 2016 Level 0             & ES                                                      \\
                          & lc1              & {\tiny{{[}String{]}}} Land cover 1                  & E20                                                     \\
                          & lc1\_label       & {\tiny{{[}String{]}}} Label of land cover           & Grassland without tree/shrub cover                      \\
                          & lu1              & {\tiny{{[}String{]}}} Land use 1                    & U111                                                    \\
                          & lu1\_label       & {\tiny{{[}String{]}}} Label of land use             & \makecell{Agriculture\\(excluding fallow land and kitchen gardens)} \\ \hline
\multirow{1}{*}{\cref{sec:data12}} & Classes          & {\tiny{{[}String{]}}} Segmentation classes & Sky, Tree, Grass, Terrain, Stonewall                    \\
\multirow{2}{*}{\cref{sec:data22}} & Quality Scores   & {\tiny{{[}Double{]}}}                               & 4.92578125                                              \\
                          & Aesthetic Scores & {\tiny{{[}Double{]}}}                               & 3.185546875                                             \\ \hline
\end{tabular}
\label{tab:lucas-csv}
\end{table}

\subsection{Data Collection}
\label{sec:data1}


\subsubsection{Image \& Basic Information}
\label{sec:data11}

We collect 1,784 scenery images (1600$\times$1200 pixels) from \cite{martinez2024semantic} for MM-LUCAS. These images were taken horizontally at the eye-height level across 1,784 different sites in 27 EU countries (see \cref{tab:27country}). The sites primarily represent agricultural and rural landscapes to ensure a diverse and comprehensive dataset. This selection aims to capture the variety and complexity of rural environments across Europe, providing a robust basis for precise analyses. Additionally, we selected specific microdata (geographical location, date, LC, LU) from \cite{martinez2024semantic}, as shown in \cref{tab:lucas-csv} (Top part).

\begin{table}[]
\caption{27 EU Countries.}
\begin{tabular}{l}
\begin{minipage}{1\linewidth}
\fontsize{7}{8}\selectfont 
\textbf{AT}: Austria, \textbf{BE}: Belgium, \textbf{BG}: Bulgaria, \textbf{CY}: Cyprus, \textbf{CZ}: Czech Republic, \textbf{DE}: Germany, \textbf{DK}: Denmark, \textbf{EE}: Estonia, \textbf{EL}: Greece, \textbf{ES}: Spain, \textbf{FI}: Finland, \textbf{FR}: France, \textbf{HR}: Croatia, \textbf{HU}: Hungary, \textbf{IE}: Ireland, \textbf{IT}: Italy, \textbf{LT}: Lithuania, \textbf{LU}: Luxembourg, \textbf{LV}: Latvia, \textbf{NL}: Netherlands, \textbf{PL}: Poland, \textbf{PT}: Portugal, \textbf{RO}: Romania, \textbf{SE}: Sweden, \textbf{SI}: Slovenia, \textbf{SK}: Slovakia, \textbf{UK}: United Kingdom
\end{minipage}
\end{tabular}
\label{tab:27country}
\end{table}

\subsubsection{Segmentation Mask}
\label{sec:data12}

Segmentation masks are provided with the same resolution as the corresponding RGB images. After thorough data cleaning, the final segmentation dataset includes 34 classes (see \cref{tab:34class}). The original segmentation masks\cite{martinez2024semantic} are single-layer grayscale images, where pixel values range from 0 to 33. To enhance intuition and visualization, we further provide color-coded segmentation masks, where each class is represented by a different color. 
\begin{table}[]
\caption{34 semantic segmentation classes (inherited from \cite{martinez2024semantic}) in MM-LUCAS.}
\begin{tabular}{l}
\begin{minipage}{1\linewidth}
\fontsize{7}{8}\selectfont 
`Sky', `Tree', `Background', `Terrain', `Plant\_Bush', `Flowerfield', `Earth\_Ground', `Mountain', `Poles', `Tower', `Automobile', `Grass', `Path', `Dense\_Woody\_Features', `Flower', `Cropfield', `Rail\_Transport', `Traffic\_Sign', `Wall', `Crop', `Fruit', `Field\_Margin', `Road', `Rock', `Orchard', `Waterbodies', `Animal', `Stonewall', `Bridge', `Lucas\_Marker', `Bark', `Person', `Well', `Building'
\end{minipage}
\end{tabular}
\label{tab:34class}
\end{table}

\subsection{Data Processing}
\label{sec:data2}

\subsubsection{Depth Estimation}
\label{sec:data21}

We adopt the advanced Depth Anything V2-Large model (335.3M parameters)\cite{yang2024depth}, which achieves robust and fine-grained depth predictions by using synthetic images, with high efficiency and accuracy.

\subsubsection{Visual Assessment}
\label{sec:data22}

We adopt Q-Align\cite{wu2023q}, which trained LLMs for visual rating by emulating human discrete-level rating processes. Compared with other similar models that rely on numerical score scaling, Q-Align aligns more with human cognition. We only evaluate image quality and aesthetics in our dataset.

\subsubsection{Landscape Question Answering (LandQA)}
\label{sec:data23}
Specifically, we utilize GPT-4o\cite{web:chatgpt4o} to construct LandQA (see \textit{Right} in \cref{fig:data-overview}) based on the LUCAS images and the corresponding annotations, land cover label (\textit{lc1\_label})\cite{martinez2024semantic}, land use label (\textit{lu1\_label})\cite{martinez2024semantic}, quality scores and aesthetic scores. The key idea is to query GPT-4o to generate diversity questions as shown in \cref{tab:landqa}.

For land cover QA and land use QA, we set up multi-choice question-answering with 4 options: \{\textit{``images:'', ``questions:'', ``options:'', ``answer:''}\}. This format can effectively evaluate the model's understanding of the land cover and land use annotations by providing distinct choices for each question.

For quality evaluation and aesthetic evaluation, we employ a basic question-answering format. This is structured as: \{\textit{images:'', questions:'', ``answer:''}\}. Note that, we design the standard of aesthetic scores as: ``\textit{0-1:Bad, 1-2:poor, 2-3: fair, 3-4: good, 4-5: excellent}''. In this setup, we suggest the model provide a straightforward answer based on the aesthetic annotations and effectively evaluate the model's ability to interpret these subjective scores.


\begin{table}[t]
\centering
\caption{Examples of GPT-4o\cite{web:chatgpt4o} generated question prompts for LandQA.}
\fontsize{7}{7}\selectfont 
\begin{tabular}{l||ll}
\toprule[1pt]
\textbf{Topics}                  & \textbf{Prompts}                                                    &  \\ \cline{1-2}
\multirow{3}{*}{Land Cover}      & What is the main type of land visible here?                         &  \\
                                 & How would you describe the surface features in this image?          &  \\
                                 & Can you specify the type of land cover in this image?               &  \\ \cline{1-2}
\multirow{3}{*}{Land Use}        & What kind of socio-economic activity is taking place in this image? &  \\
                                 & How is the land used in this area?                                  &  \\
                                 & Can you identify the main use of this land?                         &  \\ \cline{1-2}
\multirow{3}{*}{Image Quality}   & Please evaluate the quality of this photo.                          &  \\
                                 & How would you describe the quality of this photo?                   &  \\
                                 & Can you give a rating on the quality of this image?                 &  \\ \cline{1-2}
\multirow{3}{*}{Image Aesthetic} & Please provide a rating for the aesthetics of this picture.         &  \\
                                 & How do you evaluate the aesthetics of this picture?                 &  \\
                                 & Can you provide your opinion on the aesthetics of this image?       &  \\ 
\bottomrule[1pt]
\end{tabular}
\label{tab:landqa}
\end{table}

\subsection{Data Analysis}
\label{sec:data3}


\begin{figure}[]
    \centering
    \begin{subfigure}[b]{0.45\textwidth}
        \centering
        \includegraphics[width=\textwidth]{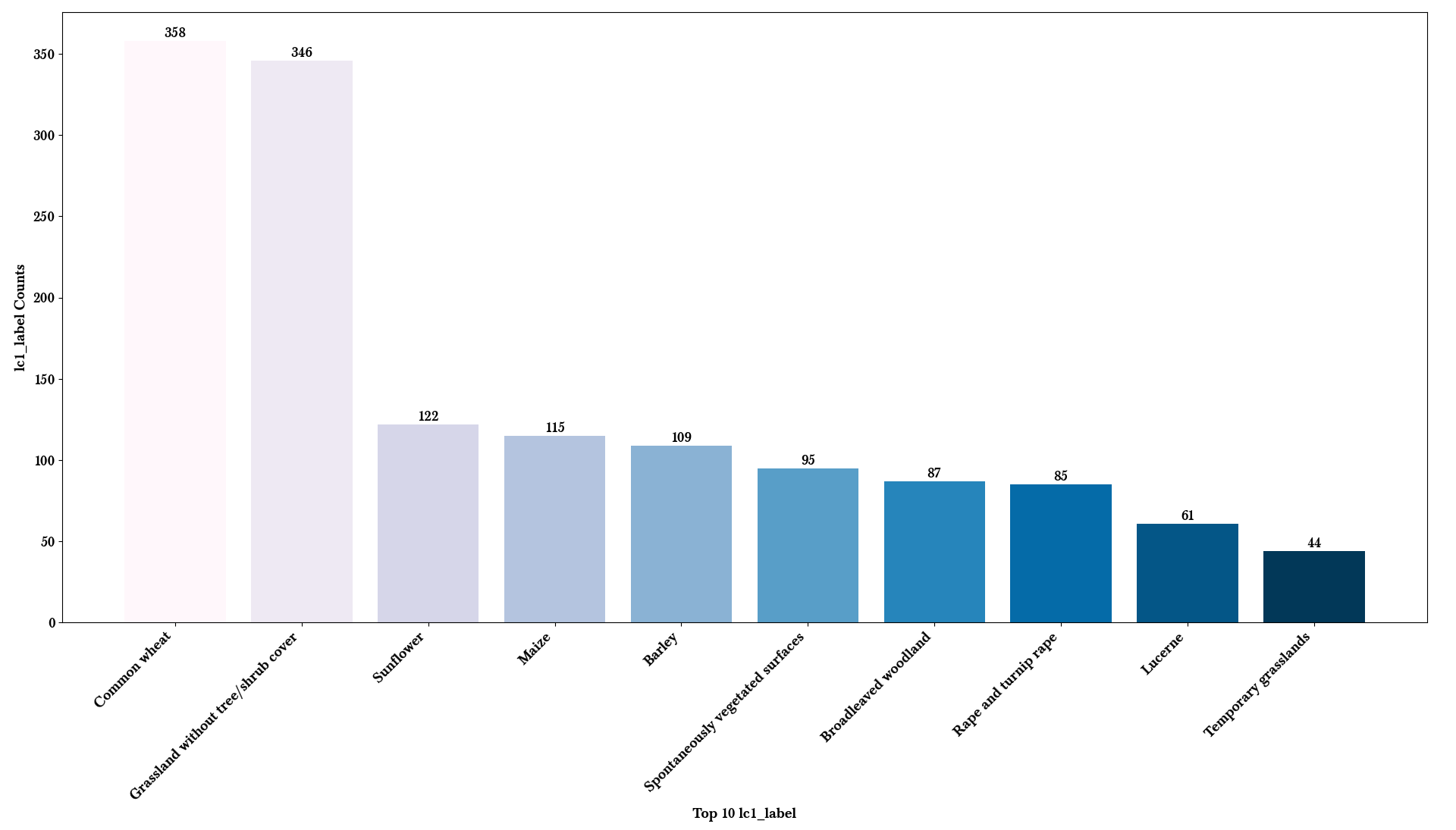}
        \caption{Top 10 Land Cover (LC) labels.}
        \label{fig:lc}
    \end{subfigure}
    \hfill
    \begin{subfigure}[b]{0.45\textwidth}
        \centering
        \includegraphics[width=\textwidth]{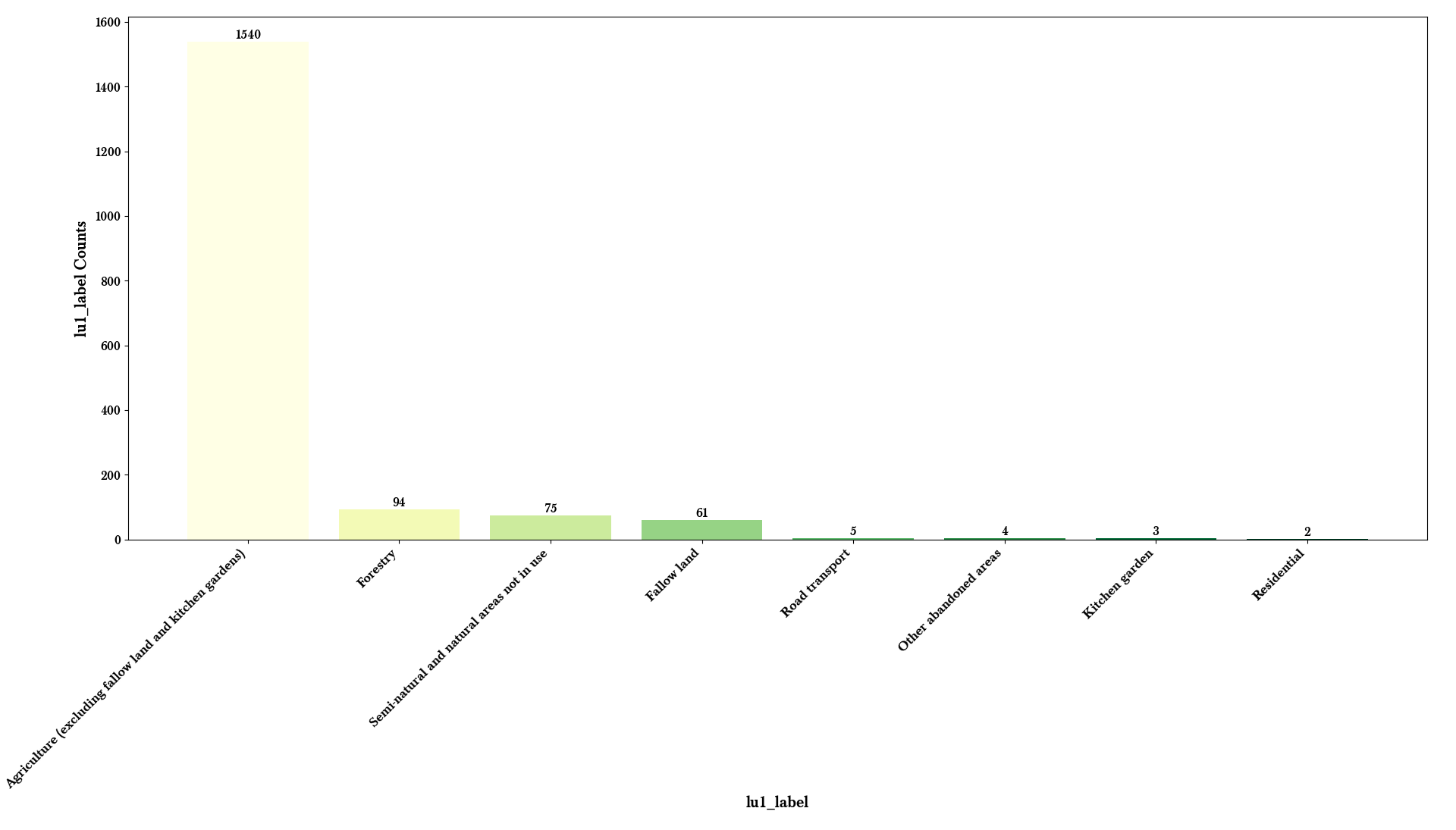}
        \caption{All Land Use (LU) labels.}
        \label{fig:lu}
    \end{subfigure}
    \caption{Distribution of the top 10 \textit{lc1\_label} and all \textit{lu1\_label} categories.}
    \label{fig:lclu}
\end{figure}

As shown in \cref{fig:lc}, we present the top 10 land cover (LC) categories in our MM-LUCAS dataset. \cref{tab:lc} list the detailed LC labels, which show a diverse range of agricultural and natural land cover types across the EU regions. \textbf{Common wheat} is the most frequently occurring. The importance of crops like \textbf{wheat, sunflower, maize, and barley} highlights the agricultural focus.

\begin{table}[]
\caption{Classification of land cover (LC) in MM-LUCAS dataset.}
\begin{tabular}{l}
\begin{minipage}{1\linewidth}
\fontsize{8}{9}\selectfont 
$\circ$ \textbf{A00}-Artificial Land: \textit{\textbf{A20}}-Artificial non-built up areas\\ 
$\circ$ \textbf{B00}-Cropland: \textit{\textbf{B10}}-Cereals, \textit{\textbf{B20}}-Root crops, \textit{\textbf{B30}}-Non-permanent industrial crops, \textit{\textbf{B40}}-Dry pulses, vegetables and flowers, \textit{\textbf{B50}}-Fodder crops, \textit{\textbf{B70}}-Permanent crops, \textit{\textbf{B80}}-Other permanent crops \\
$\circ$ \textbf{C00}-Woodland: \textit{\textbf{C10}}-Broadleaved woodland, \textit{\textbf{C20}}-Coniferous woodland, \textit{\textbf{C30}}-Mixed woodland\\
$\circ$ \textbf{D00}-Shrubland: \textit{\textbf{D10}}-Shrubland with sparse tree cover, \textit{\textbf{D20}}-Shrubland without tree cover\\ 
$\circ$ \textbf{E00}-Grassland: \textit{\textbf{E10}}-Grassland with sparse tree/shrub cover, \textit{\textbf{E20}}-Grassland without tree/shrub cover, \textit{\textbf{E30}}-Spontaneously re-vegetated surfaces\\
$\circ$ \textbf{F00}-Bare land and lichens/moss: \textit{\textbf{F10}}-Rocks and stones, \textit{\textbf{F20}}-Sand, \textit{\textbf{F40}}-Other bare soil\\ 
$\circ$ \textbf{H00}-Wetlands: \textit{\textbf{H10}}-Inland wetlands, \textit{\textbf{H20}}-Coastal wetlands
\end{minipage}
\end{tabular}
\label{tab:lc}
\end{table}

As shown in \cref{fig:lu}, we present all land use (LU) categories in our MM-LUCAS dataset, which indicate \textbf{Agriculture (excluding fallow land and kitchen gardens)}, overwhelmingly the most prevalent. \textbf{Forestry} is the second most common LU that is significantly less frequent than agriculture. \cref{tab:lu} list the detailed LU labels, indicating that although agriculture is prevailing, there is still a variety of land use types represented, including forestry, natural areas, and fallow land, providing a diverse view of the landscape.

\begin{table}[]
\caption{Classification of land use (LU) in MM-LUCAS dataset.}
\begin{tabular}{l}
\begin{minipage}{1\linewidth}
\fontsize{8}{9}\selectfont 
$\bullet$ \textbf{U100}-Primary sector: \textit{\textbf{U110}}-Agriculture, \textit{\textbf{U120}}-Forestry\\
$\bullet$ \textbf{U300}-Tertiary sector, transport, utilities \& residential: \textit{\textbf{U310}}-Transport, communication networks, storage, protection works, \textit{\textbf{U370}}-Residential\\
$\bullet$ \textbf{U400}-Unused and abandoned areas: \textit{\textbf{U110}}-Abandoned areas, \textit{\textbf{U120}}-Semi-natural and natural areas not in use
\end{minipage}
\end{tabular}
\label{tab:lu}
\end{table}

To discover how various factors influence the perceived quality and visual appeal of agriculture, we analyze the relationship between the microdata\cite{martinez2024semantic} (segmentation classes, geographical information and date) and quality scores (QS) / aesthetic scores (AS). This analysis provides insights that can help optimize agricultural practices, improve crop management, and enhance marketability.

\begin{figure}[t]
    \centering
    \includegraphics[width=0.88\linewidth]{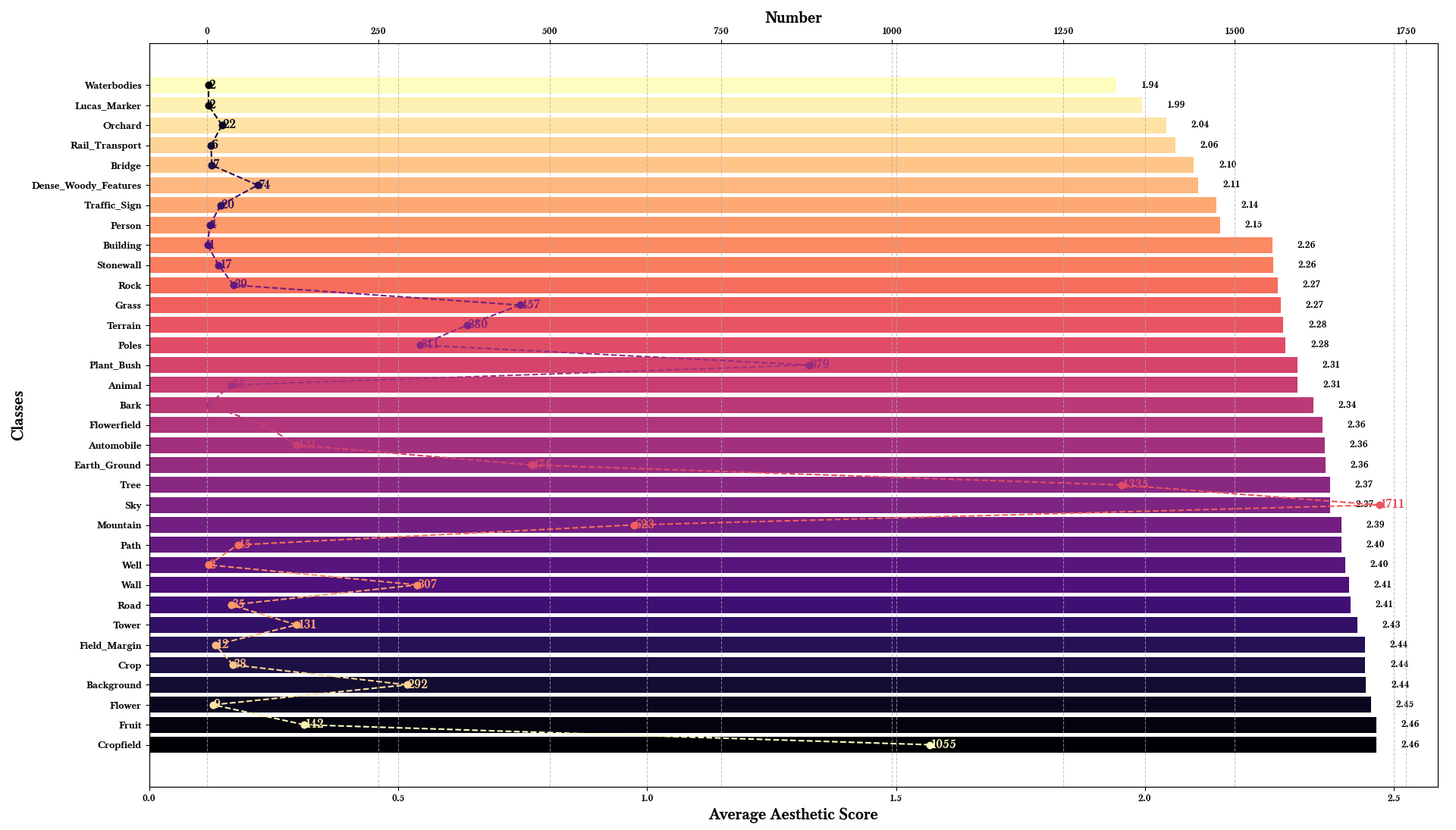}
    \caption{Average aesthetic score (AS) for different segmentation classes.}
    \label{fig:class_as}
\end{figure}

\Cref{fig:class_as} presents the relationship between the number of different segmentation classes (see \cref{tab:34class}) and their average AS. Despite the \textbf{Orchard} is infrequent, have a relatively high average AS (2.04). Natural elements, like \textbf{flower and cropfiled} have high AS (2.45 and 2.46), showing the importance of contributing to the scenery beauty.

\begin{figure}
    \centering
    \includegraphics[width=0.89\linewidth]{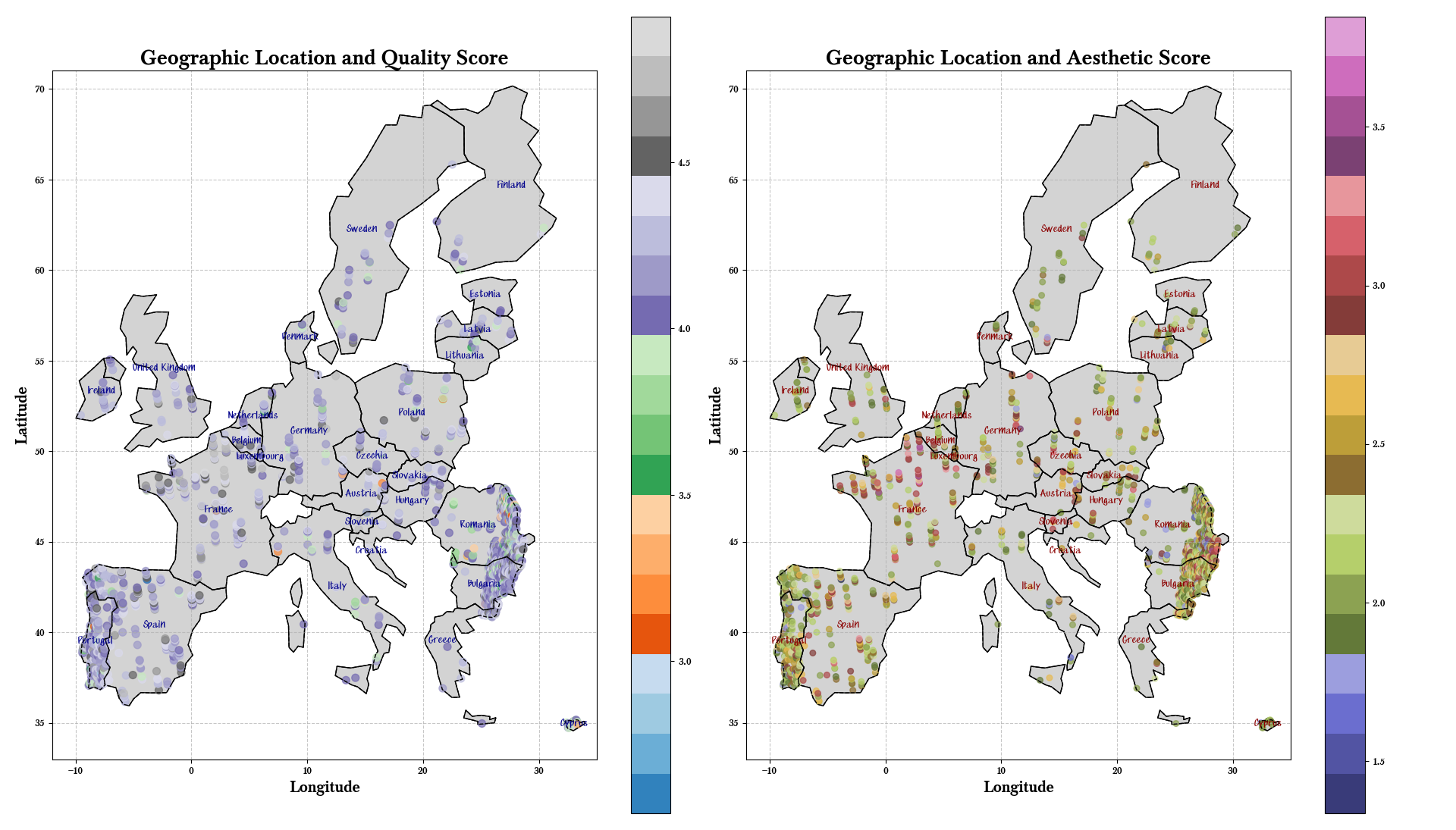}
    \caption{Spatial distribution of quality score and aesthetic score across EU.}
    \label{fig:gps_qs_as}
\end{figure}

\Cref{fig:gps_qs_as} illustrates the significant regional variability of QS and AS across the EU, which helps understand regional strengths and weaknesses, guiding local farmers and agricultural policymakers to focus on best practices that lead to higher quality and more aesthetically pleasing crops. 
\textbf{France, Germany, and the United Kingdom} have higher QS. \textbf{Romania and Bulgaria} have a mix of high and low-quality images, which could be due to diverse geographical features and environmental conditions, as shown in \cref{fig:gps_qs_as} (left). \textbf{France and parts of Romania and Bulgaria} have higher AS, while \textbf{Finland, Estonia, and Lithuania} have lower AS, as shown in \cref{fig:gps_qs_as} (right).

\begin{figure}[t]
    \centering
    \includegraphics[width=\linewidth]{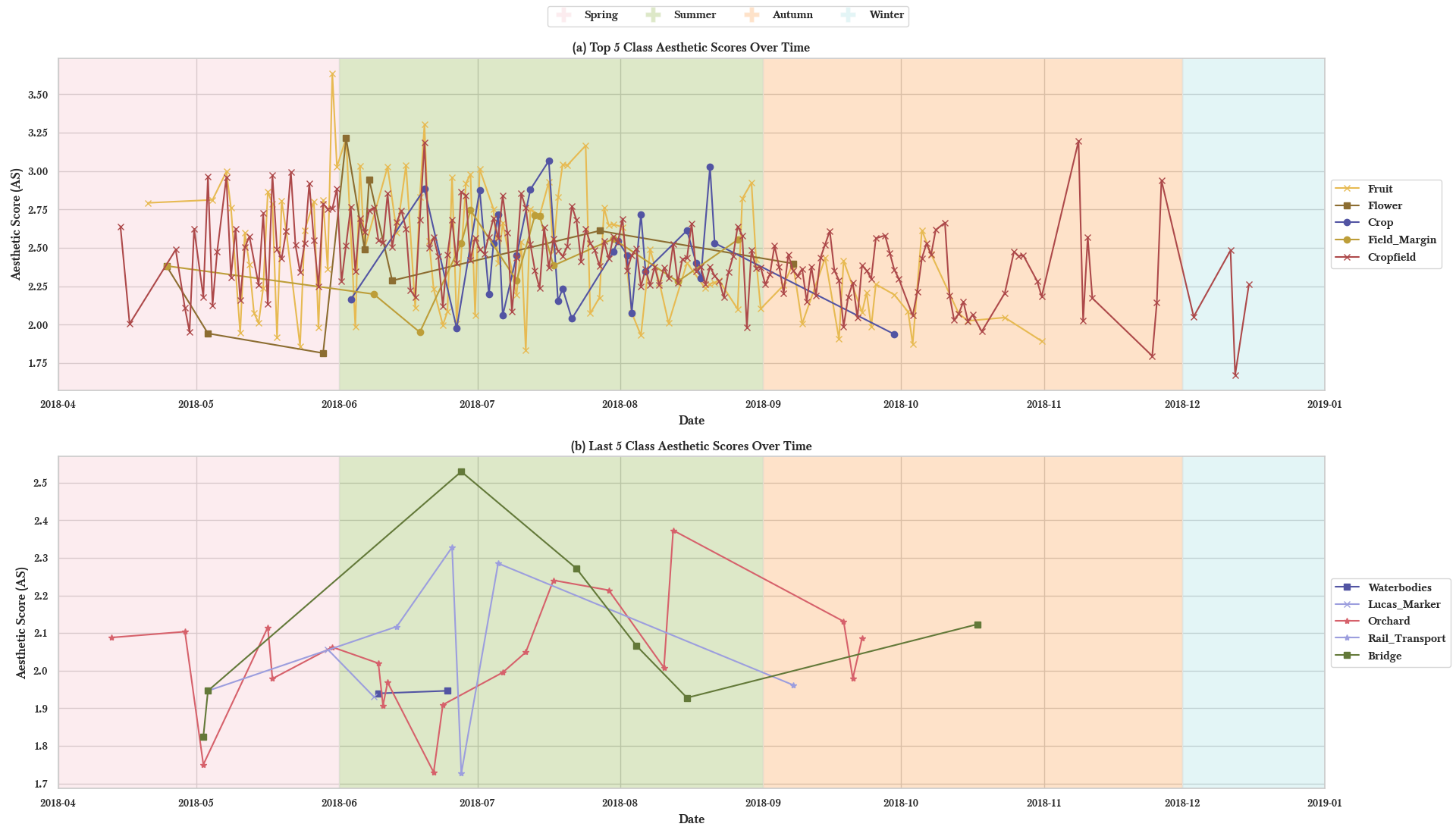}
    \caption{Seasonal changes in aesthetic score (AS) for the five most frequent classes and five least frequent classes. Zoom in for details.}
    \label{fig:class_as_date}
\end{figure}

We further analyze seasonal changes in AS for the five most frequent and five least frequent classes, which provides insights into how different agricultural products and related surrounding environments are affected by seasonal changes. As demonstrated in \cref{fig:class_as_date}(a), the top 5 classes are highly influenced by natural cycles of growth and harvest, with \textbf{spring and summer} generally showing higher and more variable scores. Strong seasonal dependency \textbf{flowers and fruit} show the highest variability, while \textbf{field margins} maintain more consistent scores. Compared to more frequent classes, the last 5 classes show lower variability in AS, indicating that their visual appeal is less correlated with seasonal changes. Overall, \textbf{summer} generally sees the highest AS due to the peak of natural beauty and crop maturity, while AS in \textbf{winter} are consistently lower across both frequent and less frequent classes. This information is necessary for optimizing planting and harvesting times, improving crop management throughout the year, and predicting market trends based on seasonality. 

\section{Experiments and Analysis}
\label{sec:exp}

\subsection{Mainstream MM-LLMs for Evaluation and Results}
\label{sec:mmllms}
We evaluate 2 open-source MM-LLMs (InterVL2-26B\cite{chen2023internvl}, and mPLUG-Owl2\cite{ye2023mplugowl2}) and 3 closed-source MM-LLMs (GPT-4o\cite{web:chatgpt4o}, Gemini-1.0 Pro\cite{web:gemini}, and Claude 3.5 Sonnet\cite{web:clause}) based on the official implementation sources in the following 2 tasks. Experimental results are shown as: ``\textbf{Level X:} \colorbox {IKKONZOME!50}{Modal} \textbf{Task} \hl{Question} Models: Results''. Some details are omitted due to page limitations.


\textbf{\hyperref[sec:l1]{Level 1:}} \colorbox {IKKONZOME!10}{T$\rightarrow$T} \textbf{Basic Question Answering} \hl{Q1: Please briefly introduce the common vegetable crops in Europe.}

$\star$ InterVL2-26B\cite{chen2023internvl}: Tomatoes, cucumbers, peppers, onions, carrots, potatoes, and lettuce...

$\star$ mPLUG-Owl2\cite{ye2023mplugowl2}: Potatoes, onions, carrots, tomatoes, peas, and lettuce. 

$\star$ Claude 3.5 Sonnet\cite{web:clause}: Potatoes; Tomatoes; Carrots; Onions; Cabbage; Lettuce; Cucumbers; Bell peppers; Peas; Beans; Broccoli and cauliflower; Spinach; Asparagus; Zucchini (courgette); Garlic...

$\star$ Gemini-1.0 Pro\cite{web:gemini}: Mass Production Leaders:
Tomatoes; Onions; Carrots ...
Other Widespread Crops:
Brassicas; Solanaceae Family ...

$\star$ GPT-4o\cite{web:chatgpt4o}: Tomatoes; Potatoes; Carrots; Onions; Cabbage; Cucumbers; Peppers (Bell and Chili); Spinach; Zucchini (Courgette) ...

\textbf{\hyperref[sec:l3]{Level 3:}} \colorbox {IKKONZOME!60}{T+I$\rightarrow$T} \textbf{Contextual Scene Analysis} \hl{Q2: Rate the aesthetics (0-5) of this image and identify key features enhancing the visual appeal.}
\begin{wrapfigure}{l}{0.4\textwidth} 
    \centering
    \includegraphics[width=0.37\textwidth]{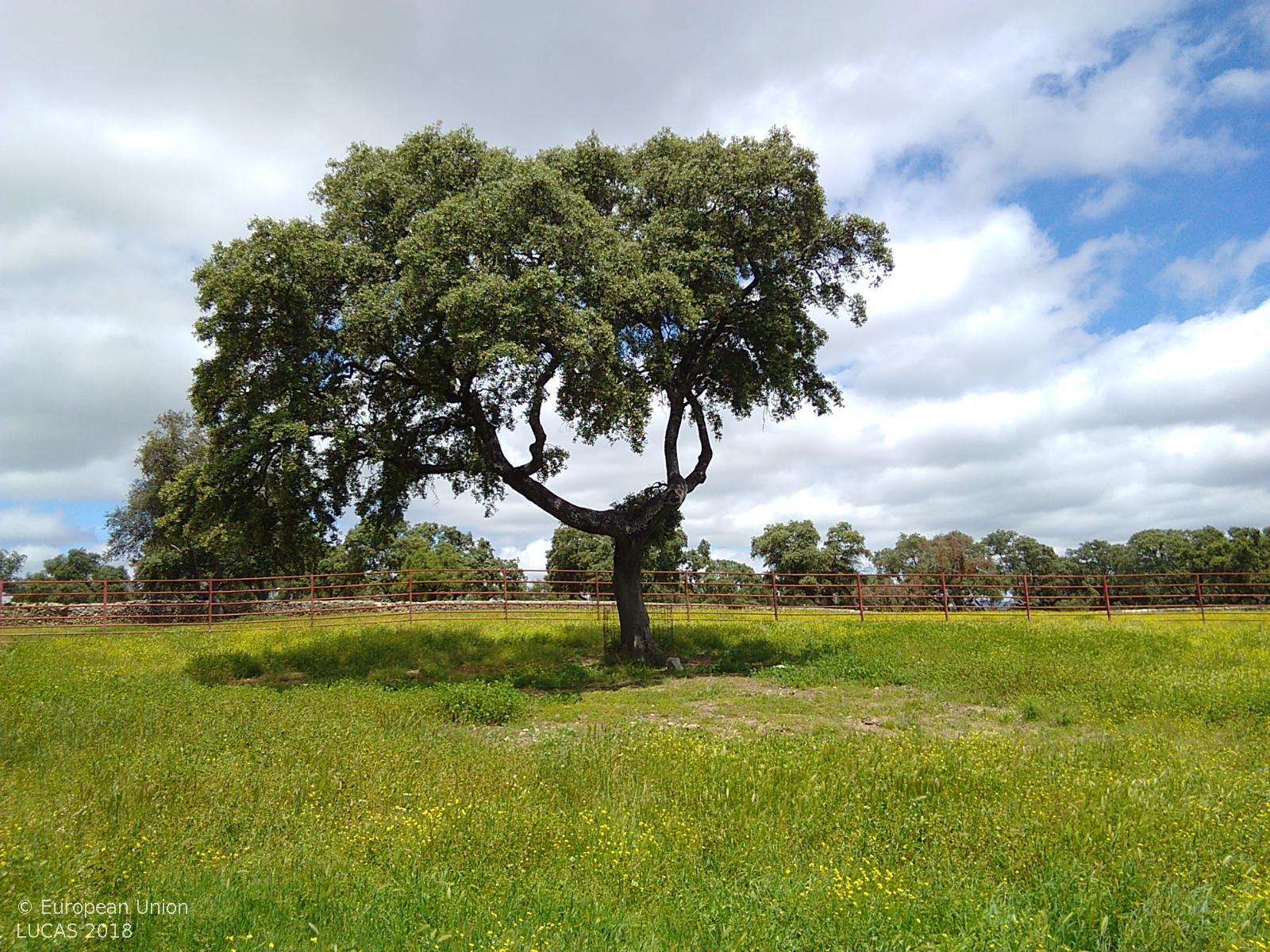} %
\end{wrapfigure}

$\star$ InterVL2-26B\cite{chen2023internvl}: \textbf{\small Score:} Composition (4/5), Lighting Conditions (4/5)$\rightarrow$4/5. \textbf{\small Key features:} Central tree; Contrast between sky \& ground.

$\star$ mPLUG-Owl2\cite{ye2023mplugowl2}: \textbf{\small Score:} 4/5. \textbf{\small Key features:} the large tree in the middle of the field, the lush green grass, and the yellow flowers surrounding the tree.

$\star$ Claude 3.5 Sonnet\cite{web:clause}: \textbf{\small Score:} 4.5/5. \textbf{\small Key features:} Focal point; Composition; Color palette; Lighting; Depth perception; Seasonal beauty; Sky drama; Natural framing; Contrast; Symmetry; Rural charm.

$\star$ Gemini-1.0 Pro\cite{web:gemini}: \textbf{\small Score:} 4/5. \textbf{\small Key features:} Simplicity and Minimalism: (4/5);
Centralized Subject: (4/5);
Contrasting Colors: (4/5);
Rule of Thirds: (4/5);
Natural Light: (4/5);
Depth of Field: (4/5).

$\star$ GPT-4o\cite{web:chatgpt4o}: \textbf{\small Score:} 4.5/5. \textbf{\small Key features:} Tree as the Focal Point; Color Contrast; Depth and Layers; Natural Symmetry and Balance; Texture and Detail; Light and Shadow; Serene Atmosphere.


\subsection{Results Analysis}
\label{sec:RA}

For \textbf{Basic Question Answering}, InterVL2-26B and mPLUG-Owl2 provide concise lists of 6-7 common vegetables, offering straightforward and relevant responses. In contrast, Claude 3.5 Sonnet and GPT-4o present more comprehensive lists, reflecting a broader knowledge. Gemini-1.0 Pro uniquely categorizes crops into mass production leaders and other widespread crops, highlighting the economic significance and scale of different crops.

For \textbf{Contextual Scene Analysis}, the five MM-LLMs consistently rate the AS between 4 and 4.5 out of 5. Key visual features include the tree, color contrast, and depth. While InterVL2-26B, mPLUG-Owl2, and Gemini-1.0 Pro focus on the tree and contrast as central elements. Claude 3.5 Sonnet and GPT-4o provide a more comprehensive analysis, especially highlighting seasonal beauty and light, offering a deeper understanding of the scene aesthetic.

\section{Discussion}
\label{sec:diss}

\,\,\,\quad \textbf{{Correlation of MM-LUCAS and AgriBench.}} Semantic segmentation masks and depth maps play an important role in AgriBench, evaluating the agricultural MM-LLM tasks by offering detailed visual and spatial information. They provide an extra condition for accurately assessing tasks such as \textit{species classification, object counting}, and \textit{dense object counting} within complex scenes. They also improve \textit{contextual scene analysis} and \textit{environmental impact prediction} by offering deeper insights into the relationships between objects and their surroundings. Additionally, they support \textit{scene projection, scene generation}, and \textit{sustainability recommendations} by ensuring realistic spatial relationships and providing key data for informed decision-making.


\textbf{Responsible AI.}
Another important challenge is the requirement for MM-LLMs to be interpretable and explainable, even in advanced models. Responsible AI considers various aspects, such as bias, fairness, transparency, human oversight, \etc, which ensures that users can understand how and why a model makes certain decisions, as this is crucial for decision-making and gaining the trust of stakeholders.  For instance, in healthcare, understanding the decision-making process of an AI model can help doctors trust and effectively use AI recommendations. Similarly, in agriculture, farmers need clear insights into AI predictions to ensure regulatory compliance and to make informed decisions. This trust ensures that AI systems are effective, trustworthy, and ethical, which is foundational for the widespread implementation and acceptance of AI technologies in various research domains and real-world applications.  

\textbf{Limitations and Future Works.} Our initial evaluation is primarily qualitative. We plan to extend our benchmark by including evaluation metrics, from perception (user satisfaction scores and expert reviews) to cognition (accuracy rate and consistency score), to provide a comprehensive evaluation that combines both qualitative and quantitative.


\section{Conclusion}
\label{sec:conclusion}

This paper introduces AgriBench, the first agriculture benchmark evaluating MM-LLMs across multi-modality and multi-tasks dimensions, supporting a broader range of agriculture scenarios and applications. We further introduce MM-LUCAS, the first multimodal agriculture dataset with multiple annotations. This initial exploration is still a work in progress and represents the first step designed to align with AgriBench. Finally, we compare 5 MM-LLMs on AgriBench and highlight significant insights in specialized domain MM-LLMs for further exploration. AgriBench represents the initial step and a significant contribution to developing agricultural MM-LLMs. We aim to boost the progress of AI technologies specifically for agricultural needs, benefiting both researchers and practitioners.

\subsubsection*{Acknowledgment.} 
This study was supported by Bundesministerium für Bildung und Forschung (BMBF) project “KI und Citizen Science gestütztes Monitoring von zertifizierten Biodiversitätsprojekten” (16LW0441).

%
%
\bibliographystyle{splncs04}
\bibliography{ref}
\end{document}